\pdfoutput=1

\documentclass[11pt]{article}

\usepackage[final]{acl}

\usepackage{times}
\usepackage{latexsym}

\usepackage{CJKutf8}
\usepackage{booktabs}
\usepackage{graphicx}
\usepackage{subfigure}
\usepackage{comment}
\usepackage{xspace}
\usepackage{adjustbox}

\usepackage[T1]{fontenc}

\usepackage[utf8]{inputenc}

\usepackage{microtype}
\usepackage{xcolor}
\usepackage{fancyhdr}
\newcommand{\INFSYS}{\textsc{InfMoE}\xspace}

\usepackage{xcolor, soul}
	\definecolor{zzycolor}{rgb}{0.688, 0.588, 0.188}

\usepackage{bbding}

\title{CPM-2: Large-scale Cost-effective Pre-trained Language Models}

\author{Zhengyan Zhang{$^*$}, Yuxian Gu{$^*$}, Xu Han{$^*$}, Shengqi Chen{$^*$}, Chaojun Xiao{$^*$}, Zhenbo Sun \\
\textbf{Yuan Yao, Fanchao Qi, Jian Guan, Pei Ke, Yanzheng Cai, Guoyang Zeng, Zhixing Tan} \\
\textbf{Zhiyuan Liu$^\dagger$, Minlie Huang$^\dagger$, Wentao Han$^\dagger$, Yang Liu, Xiaoyan Zhu, Maosong Sun} \\
Department of Computer Science and Technology, Tsinghua University \& BAAI
}

\begin{document}
\maketitle

\begin{CJK*}{UTF8}{gbsn}

\begin{abstract}
In recent years, the size of pre-trained language models (PLMs) has grown by leaps and bounds. However, efficiency issues of these large-scale PLMs limit their utilization in real-world scenarios. We present a suite of cost-effective techniques for the use of PLMs to deal with the efficiency issues of pre-training, fine-tuning, and inference. (1) We introduce knowledge inheritance to accelerate the pre-training process by exploiting existing PLMs instead of training models from scratch. (2) We explore the best practice of prompt tuning with large-scale PLMs. Compared with conventional fine-tuning, prompt tuning significantly reduces the number of task-specific parameters. (3) We implement a new inference toolkit, namely \INFSYS, for using large-scale PLMs with limited computational resources. Based on our cost-effective pipeline, we pre-train two models: an encoder-decoder bilingual model with $11$ billion parameters (CPM-2) and its corresponding MoE version with $198$ billion parameters. In our experiments, we compare CPM-2 with mT5 on downstream tasks. Experimental results show that CPM-2 has excellent general language intelligence. Moreover, we validate the efficiency of \INFSYS when conducting inference of large-scale models having tens of billions of parameters on a single GPU. All source code and model parameters are available at \url{https://github.com/TsinghuaAI/CPM}.
\end{abstract}

\section{Introduction}

{\let\thefootnote\relax\footnotetext{$^*$ Equal contribution}}
{\let\thefootnote\relax\footnotetext{$^\dagger$ Corresponding authors: Z. Liu (liuzy@tsinghua.edu.cn), M. Huang (aihuang@tsinghua.edu.cn), W. Han (hanwentao@tsinghua.edu.cn)}}

Training much larger models is an important research direction in deep learning~\citep{DBLP:conf/slsp/Bengio13}. Recently, pre-training has become the mainstream technique to develop large-scale neural networks and achieved great success in both computer vision (CV) and natural language processing (NLP)~\citep{ResNet,ViT,BERT}. Especially, there are some much larger pre-trained language models (PLMs) with hundreds of billions of parameters, such as GPT-3~\cite{brown2020language}, PANGU-$\alpha$~\cite{zeng2021pangu}, and Switch-Transformer~\cite{fedus2021switch}.

However, the cost of using PLMs is increasing rapidly with the growth of model sizes and becomes unaffordable for most users and researchers. The cost consists of three parts. (1)~\textbf{Large computation cost} for pre-training: a super large model requires several weeks of pre-training with thousands of GPUs. (2)~\textbf{Large storage cost} for fine-tuned models: a super large model usually takes hundreds of gigabytes (GBs) to store, and we need to store as many models as downstream tasks. (3)~\textbf{Strict equipment requirement} for inference: it is common to use multiple GPUs for the inference of a super large model, so these models are hard to be used with limited computation resources.

\begin{table*}[t]
    \centering
    \small
    \begin{adjustbox}{width=\linewidth,center}
    \begin{tabular}{lrrrrrrrrr}
    \toprule
               & $n_{param}$ & $L$  & $n_{head}$ & $d_{head}$ & $d_{ff}$ & $d_{model}$ &  Encoder & Decoder & MoE  \\
    \midrule
    CPM-Small  & 109M     & 12 & 12      & 64      & 3,072  & 768      & \XSolidBrush  & \Checkmark & \XSolidBrush \\
    CPM-Medium & 334M     & 24 & 16      & 64      & 4,096  & 1,024     & \XSolidBrush  & \Checkmark &    \XSolidBrush \\
    CPM-Large  & 2.6B     & 32 & 32      & 80      & 10,240 & 2,560     & \XSolidBrush  & \Checkmark &    \XSolidBrush\\
    \midrule
    CPM-2      & 11B      & 24 & 64      & 64      & 10,240 & 4,096     & \Checkmark  & \Checkmark &  \XSolidBrush\\
    CPM-2-MoE  & 198B     & 24 & 64      & 64      & 10,240 & 4,096     & \Checkmark  & \Checkmark &  \Checkmark\\
    \bottomrule
    \end{tabular}
    \end{adjustbox}
    \caption{Comparisons between CPM and CPM-2. $n_{param}$ is the amount of model parameters. $L$ is the number of model layers. $n_{head}$ is the number of attention heads in each layer. $d_{head}$ is the dimension of each attention head. $d_{ff}$ is the intermediate dimension of feed-forward layers. $d_{model}$ is the dimension of hidden states.}
    \label{tab:sizes}
    \end{table*}

To reduce the cost of large-scale PLMs from its pre-training to fine-tuning, we try to improve the whole pipeline of developing PLMs as follows:

(1) We adopt knowledge inheritance~\citep{qin2021ki} to accelerate the pre-training process. Current PLMs are usually trained from scratch on pre-training data via self-supervised methods, while there exist many PLMs that can also provide much knowledge. Knowledge inheritance aims to use the knowledge of existing PLMs to help the pre-training of new models.

(2) We use prompt tuning~\citep{prompt-tuning} instead of fine-tuning to reduce the storage of task-specific parameters. With prompt tuning, we only need to save the embeddings of prompt tokens, whose parameters are usually less than $0.01\%$ of the whole model parameters.

(3) We design a high-performance and memory-efficient inference framework \INFSYS with a dynamically-scheduled offloading strategy, to support the inference of MoE models on a single GPU.

Based on our optimized pipeline for PLMs, we develop two large-scale \textbf{C}ost-efficient \textbf{P}re-trained language \textbf{M}odels (CPM-2), an Chinese-English bilingual models with $11$ billion parameters and its Mixture-of-Experts (MoE) version with $198$ billion parameters. Specifically, we accelerate the pre-training process by dividing the pre-training process into three stages with knowledge inheritance: Chinese pre-training, bilingual pre-training, and MoE pre-training. Then, we compare CPM-2 with mT5~\cite{xue2020mt5}. Experimental results show that CPM-2 has excellent general language intelligence, including seven specific language capabilities. Based on CPM-2, we search for the best practice of prompt tuning. We find that (1) the positions of prompts are crucial and (2) combining prompt tuning and fine-tuning can lead to better results. Finally, we introduce \INFSYS for users to conduct inference of large-scale models with tens of billions of parameters on a single GPU.

\section{Pre-Training}

In this section, we present the pre-training details of CPM-2.

\subsection{Model}

To reach a good balance between language understanding and generation, we develop CPM-2 based on a standard Transformer architecture consisting of a bidirectional encoder and a unidirectional decoder~\citep{DBLP:conf/nips/VaswaniSPUJGKP17}. Correspondingly, we adopt a variant of Masked Language Model (MLM)~\citep{BERT,T5}, which is designed for encoder-decoder models. We construct the encoder input by randomly replacing several spans with different special tokens, and then ask the decoder to predict the replaced spans in turn. For example, given the original input, ``These are issues which future studies may seek to address'', we can construct the encoder input, ``These are {\ttfamily[X]} which {\ttfamily[Y]} may seek to address'', and the decoder target output ``{\ttfamily[X]} issues {\ttfamily[Y]} future studies {\ttfamily[Z]}''. {\ttfamily[X]}, {\ttfamily[Y]}, {\ttfamily[Z]} are special tokens, where {\ttfamily[X]} and {\ttfamily[Y]} are used to represent different spans and {\ttfamily[Z]} is used to represent the end of the output. Note that the ratio between the replaced tokens and the total tokens is 15\% and the average length of replaced spans is set to 10.

The comparisons between our models and CPM~\cite{zhang2020cpm} are presented in Table~\ref{tab:sizes}. To efficiently store model parameters on GPUs, we use the model parallelism~\citep{megatron}, which splits self-attention layers and feed-forward layers along the width dimension, and finally distributes the partitions of one model on 4 GPUs.

To reduce memory requirements and speed up pre-training, we use mixed-precision training~\citep{micikevicius2018mixed}, gradient checkpointing~\citep{chen2016training} and ZERO-stage-1 optimization~\citep{rajbhandari2020zero,rasley2020deepspeed}. 

For CPM-2-MoE, we expand the feed-forward layer of each Transformer block to multiple experts. During the forward pass, for each token, we select one expert according to its current hidden state with a gating function. We balance the expert selection using the planning approach of BASE Layers~\citep{lewis2021base}. Mixture-of-experts is an important technique for large-scale models because it can significantly improve the model capacity without extra computation cost~\citep{MoE,lepikhin2020gshard,fedus2021switch}.

\begin{table*}[t]
    \centering
    \footnotesize
    \begin{tabular}{lrrrrrrrr}
    \toprule
          & CCPM &  $\textrm{C}^3$   & Sogou-Log & WMT20-enzh  & Math23K  & LCSTS & LCQMC & AdGen \\ \midrule
    Train &   21k  &   8k  & 8,052k & 21,000k     &  21k     & 2,400k  &  238k  &  114k    \\
    Valid &  2.7k  & 2.7k   &  500k &    2k     &   1k     &   8.6k  & 8.8k  &    1k \\
    Test  &  2.7k  & 2.7k  & 1k  &    2k     &   1k     &   0.7k   & 12.5k  &    3k \\
    \bottomrule
    \end{tabular}
    \caption{Numbers of instances in each dataset.}
    \vspace{-1em}
    \label{tab:datasets}
    \end{table*}

\subsection{Data Processing}
We pre-train our model on WuDaoCorpus~\citep{WudaoCorpora}, which contains 2.3TB cleaned Chinese data as well as 300GB cleaned English data. Data in both languages are collected from multiple domains, including encyclopedia, novels, Q\&A, scientific literature, e-book, news, and reviews.

To efficiently tokenize our pre-training corpus, we explore to reduce the redundancy brought by sentencepiece~\cite{DBLP:conf/emnlp/KudoR18} to improve the vocabulary of CPM.

We find that the original sentencepiece tokenizer will insert many redundant white space tokens "\_" to tokenized sequences. This makes the sequences become much longer. Since the implementation of sentencepiece has a weak encapsulation of interfaces, it is unfriendly towards programmers. Inspired by WoBERT~\citep{zhuiyiwobert}, we replace the sentencepiece tokenizer with a simple prefix matching and remove the white space insertion. Compared with sentencepiece, our newly-implemented tokenizer is more effective and easier to use.

Besides, in the writing system of Chinese, it is not important whether a token in the vocabulary appears at the beginning of a word or not, we merge the tokens like ``快乐'' (happy) and ``\_快乐'' (\_happy) to a single token ``快乐'' (happy) to simplify the vocabulary. 


\subsection{Pre-Training with Knowledge Inheritance}
\label{sec:pretraining}

The pre-training process of CPM-2 can be divided into three stages: Chinese pre-training, bilingual pre-training, and MoE pre-training. Compared to training models from scratch, multi-stage training with knowledge inheritance~\cite{qin2021ki} can significantly reduce the computation cost.

\textbf{Chinese Stage.} In this stage, we only use Chinese texts as the training data. We suppose the model can focus on learning Chinese information and have a good basis to generalize to other languages.

\textbf{Bilingual Stage.} In this stage, we further pre-train the model from the Chinese stage on both Chinese and English texts. There are two main challenges, how to initialize the input embeddings of English tokens and how to prevent the model from catastrophic forgetting. (1) When initializing English embeddings, we use the embeddings of their prefixes to initialize their embeddings, making the English tokens more familiar to the model. If all prefixes of an English token are not in the original vocabulary, we randomly select an existing token embedding for initialization. (2) To eliminate the effect of catastrophic forgetting, we carefully design the ratio between English data and Chinese data. In the experiment, we find 1:2 can well maintain the language knowledge of Chinese and capture new knowledge of English.

\textbf{MoE Stage.} In this stage, we duplicate the model from the bilingual stage several times to initialize an MoE model. For the gating network, we adopt a random projection as a local sensitive hashing function~\citep{DBLP:journals/toc/Har-PeledIM12} and will not update the gating network in this stage. We suppose that the representation space of the model of the second stage is well organized, where similar tokens should use the same expert.

\section{Evaluation Setups}

To validate the effectiveness of our model, we evaluate CPM-2 on a general language intelligence benchmark, CUGE~\cite{yao2021cuge}. CUGE consists of $40$ mainstream Chinese NLP datasets and each dataset is categorized into one of the important types of language capabilities. Due to the limitation of computation, we select a representative dataset for each language capability to speed up the experiments. We describe each language capability and dataset as follows. The detailed statistics of these datasets are shown in Table~\ref{tab:datasets}.

\textbf{Recall Capability.} Recall capability aims to evaluate the models' ability to memorize and apply the general literature knowledge, such as the famous quotes, classical poems, and idioms. We adopt Chinese Classical Poetry Matching Dataset~(CCPM)~\citep{li2021CCPM} to test the models' recall ability. Given a modern Chinese translation of a classic poem, the model is required to select the corresponding poem from four candidates.

\begin{table*}[t]
	\begin{adjustbox}{width=\textwidth}
	\small
	\centering
	\begin{tabular}{l|rrrrrr|rr|r}
	\toprule
		  & \multicolumn{1}{c}{CCPM}  & \multicolumn{1}{c}{$\textrm{C}^3$} & \multicolumn{1}{c}{Sogou-Log} & \multicolumn{1}{c}{WMT20} & \multicolumn{1}{c}{Math23K}         & \multicolumn{1}{c|}{LCSTS}  & \multicolumn{1}{c}{LCQMC} & \multicolumn{1}{c|}{AdGen} & \multicolumn{1}{c}{CUGE}\\
		\midrule
		  & \multicolumn{1}{c}{Acc} & \multicolumn{1}{c}{Acc}    & \multicolumn{1}{c}{MRR/NDCG} &\multicolumn{1}{c}{BLEU}    & \multicolumn{1}{c}{Acc}  & \multicolumn{1}{c|}{Rouge-L}     & \multicolumn{1}{c}{Acc} & \multicolumn{1}{c|}{BLEU/Distinct} & \multicolumn{1}{c}{Score}\\
	mT5-small   &  87.7$_{(100)}$      &  41.5$_{(100)}$    &   29.2/29.2$_{(100)}$       &  9.1$_{(100)}$  &  18.4$_{(100)}$       &  33.1$_{(100)}$  &  82.1$_{(100)}$  &   10.2/32.3$_{(100)}$ & \multicolumn{1}{c}{100}\\
	mT5-large   &    89.9$_{(102)}$     &  56.3$_{(136)}$   &   32.2/31.1$_{(108)}$ &   11.1$_{(122)}$       &  34.3$_{(186)}$  & 34.4$_{(104)}$    &    85.0$_{(104)}$    &  10.0/35.5$_{(104)}$ & \multicolumn{1}{c}{126} \\
	mT5-XXL   &    90.6$_{(103)}$     & \textbf{86.4}$_{(\mathbf{208})}$     & \textbf{36.9}/34.9$_{(\mathbf{123})}$ &  24.0$_{(264)}$       & 61.6$_{(335)}$       &    34.8$_{(105)}$    &  88.3$_{(108)}$ & 9.8/68.7$_{(154)}$ & \multicolumn{1}{c}{190}\\
	CPM-2 &     \textbf{91.6}$_{(\mathbf{104})}$    & 86.1$_{(207)}$   & 36.3/\textbf{35.5}$_{(\mathbf{123})}$ & \textbf{26.2}$_{(\mathbf{288})}$         & \textbf{69.4}$_{(\mathbf{377})}$     &   \textbf{35.9}$_{(\mathbf{108})}$  &  \textbf{89.2}$_{(\mathbf{109})}$ & \textbf{10.6/70.2}$_{(\mathbf{161})}$ & \multicolumn{1}{c}{\textbf{198}}\\ \bottomrule
	\end{tabular}
	\end{adjustbox}
	\caption{Performance of mT5 and CPM-2 with fine-tuning. We use the first 6 datasets, which makes up the lite version of CUGE, to compute the overall CUGE scores (\%). The numbers in brackets are the CUEG scores (\%) for each dataset.}
	\label{tab:full_param}
	\end{table*}

\textbf{Comprehension Capability.} Comprehension capability aims to evaluate the models' ability to understand the given text and perform reasoning for specific tasks. For this capability, we select the C$^3$ dataset~\cite{sun2020investigating} to evaluate our model. C$^3$ is a free-form multiple-choice reading comprehension dataset, which requires the model to understand the given documents or dialogues and answer several related questions.


\textbf{Calculation Capability.} Calculation capability aims to test the models' ability to perform numerical reasoning. For this capability, we select Math23K~\cite{wang2017deep}, which consists of tens of thousands of real math word problems for elementary school students.

\textbf{Cross-lingual Capability.} Cross-lingual capability aims to evaluate the models' performance in understanding multi-lingual text. We adopt the machine translation task to evaluate the ability of CPM-2 in understanding English and Chinese sentences. The dataset we used in this task is provided by WMT20~\citep{barrault-etal-2020-findings}.

\textbf{Summarization Capability.} Summarization requires the model to read a long document and produce a concise summary while keeping the key information. We utilize LCSTS~\cite{hu2015lcsts} to evaluate the summarization capability. LCSTS consists of tweets and their corresponding abstracts from the largest Chinese microblogging website~(Sina Weibo).

\textbf{Classification Capability.} Text classification is a classic task in natural language processing. We evaluate the classification capability with a large-scale natural language inference dataset, LCQMC~\cite{liu-etal-2018-lcqmc}. Given two questions, LCQMC requires the model to answer whether the two questions express similar intent.

\textbf{Generation Capability.} Text generation is one of the important tasks in natural language processing, which aims to generate fluent and diverse text. We adopt the AdGen~\cite{shao2019long} as our benchmark, which requires the model to generate long advertising text given the several keywords.


We transform different tasks to a unified sequence-to-sequence format except for Sogou-Log. For Sogou-log, we train models in a contrastive manner following previous work~\cite{DBLP:conf/acl/SunLXL18}.
Besides the original metrics, such as accuracy and BLEU, we also report the CUGE score of each dataset, which is the percentage between the performance of the evaluated model and that of mT5-small.

We compare our model with mT5~\cite{xue2020mt5}, including mT5-small, mtT5-large, and mT5-XXL. Notably, mT5-XXL also adopts an encoder-decoder architecture with $13$ billion parameters, which is comparable to CPM-2. To the best of our knowledge, Pangu-$\alpha$~\cite{zeng2021pangu} with $200$ billion parameters is the largest Chinese pre-trained language model, which performs well in many downstream tasks. However, the parameters of Pangu-$\alpha$ are not publicly available, and thus we leave the comparison between CPM-2 and Pangu-$\alpha$ as future work. 

\section{Fine-Tuning} \label{sec:finetune}

In this section, we fine-tune CPM-2 and mT5 on downstream tasks to evaluate their general language intelligence.

\subsection{Experimental Setups}

We adjust maximum lengths, batch sizes, learning rates for different models and datasets. Considering that the tokenizers of CPM-2 and mT5 are different, we first tokenize the whole dataset and then set the maximum length of samples as the maximum length instead of a pre-defined length. For the batch size, we search from 128  to 512 to ensure the number of input tokens is around $2^{16}$ following \citet{T5}. For learning rates, we search from 1e-6 to 1e-4 and we find that larger models prefer smaller values.

\subsection{Results}

\begin{table*}[t]
    \small
    \centering
    \begin{tabular}{lcccccccc}
    \toprule
          & \multicolumn{1}{c}{CCPM}  & \multicolumn{1}{c}{$\textrm{C}^3$} & Sogou-Log & \multicolumn{1}{c}{WMT20} & \multicolumn{1}{c}{Math23K} & \multicolumn{1}{c}{LCSTS} & \multicolumn{1}{c}{LCQMC} & \multicolumn{1}{c}{AdGen} \\ \midrule
    \multicolumn{9}{c}{Performance on test set} \\ \midrule
    & Acc   & Acc & MRR/NDCG     & BLEU    & Acc      & Rouge-L & Acc & BLEU/Distinct \\
    CPM-2-F                          & 91.63 & 86.05  & 36.28/35.49 & 26.21  & 69.37     & 35.88  & 89.16 & 10.60/70.22  \\
    CPM-2-P                          & 90.85 & 85.33 & 30.28/30.64 &  24.13  & 67.48       & 34.17 & 88.36  & \ \ 8.63/72.02 \\
    $\Delta(\textrm{P}-\textrm{F})$  & \,-0.78 & \,-0.72 & \,-6.00/\,-4.85 &  \,-2.08  & \,-1.89      & \,-1.71 & \,-0.80 & \,-1.97/+1.80   \\
    \midrule
    \multicolumn{9}{c}{GPU memory usage(\%)} \\ \midrule
    CPM-2-F       &   98    & 96  & 88 & 98   &    93  &    98      & 98       &   98     \\
    CPM-2-P       & 50      & 46    &  49 &   75   &   68            &    76 & 54   & 53  \\
    $\Delta(\textrm{P}-\textrm{F})$ & -48  & -50  & -39  & -23  &   -25          &  -22   & -44  & -45   \\
    \bottomrule
    \end{tabular}
    \caption{Comparisons between fine-tuning and prompt tuning. CPM-2-F represents fine-tuning. CPM-2-P represents prompt tuning. $\Delta(\textrm{P}-\textrm{F})$ means the difference between fine-tuning and prompt tuning.} \label{tab:prompt_exp}
    \end{table*}

The results of fine-tuning are shown in Table~\ref{tab:full_param}. We observe that CPM-2 is better than mT5 in most language capabilities, including Chinese language understanding, generation and English to Chinese translation. Especially, CPM-2 outperforms mT5-XXL by over 10\% in Math23K, which is for calculation capability. On the overall CUGE score, CPM-2 outperforms mT5-XXL by over 4\%. This demonstrates that CPM-2 is an omnipotent large-scale multi-lingual PLM.

\section{Prompt Tuning}
\label{sec:prompt}

In this section, we study \textbf{prompt tuning}~\cite{prompt-tuning,DBLP:conf/naacl/QinE21,DBLP:journals/corr/abs-2101-00190,DBLP:journals/corr/abs-2103-10385,DBLP:journals/corr/abs-2101-00121} based on CPM-2. Different from conventional fine-tuning, prompt tuning inserts several prompt tokens into the original inputs and only updates the parameters of the inserted prompt tokens. For better clarification, we refer to the conventional full-parameter fine-tuning~\cite{BERT} as \textbf{full-model tuning}. Throughout our experiments, we keep the number of prompt tokens as $100$ to control the number of trainable parameters and initialize the parameters randomly. In the prompt tuning setting, the amount of the parameters needed to update is only $409.6$K. Compared to the $11$B parameters of full-model tuning, prompt tuning only needs to modify $0.0037\%$ parameters. We present the main results of prompt tuning in Section~\ref{sec:prompt_main}. We also explore how the positions of inserted prompt tokens affect the model performance (Section~\ref{sec:prompt_pos}), how the prompt tokens work (Section~\ref{sec:prompt_work}), and propose a two-stage fine-tuning strategy to improve model performance on downstream tasks (Section~\ref{sec:two_stage}).

\subsection{Main Results} \label{sec:prompt_main}
We present the model performance and GPU memory usage of both full-model tuning and prompt tuning in Table~\ref{tab:prompt_exp}. From the results, we have two observations. (1) With prompt tuning, CPM-2 can achieve comparable performance to full-model tuning on most of tasks. However, prompt tuning significantly degrades the performance on Sogou-Log. The reason may be that Sogou-Log adopts a contrastive loss, which is different from other datasets and difficult to optimize under prompt tuning. (2) Prompt tuning is much more memory-efficient. The results of the GPU memory usage show that prompt tuning can save at most $50\%$ GPU memory compared with full-model tuning. This is because when the model is trained with the Adam optimizer, gradients and optimizer states account for a large proportion of the overall GPU memory. Since the number of parameters needed to be optimized is much smaller in prompt tuning, the total sizes of gradient tensors and optimizer state tensors decrease. Note that small sizes of gradient tensors and optimizer state tensors also lead to small communication overhead during the synchronization of distributed training. This makes the optimization of a single step in prompt tuning faster than full-model tuning. However, we also observe that it takes much more steps for prompt tuning to converge than full-model tuning, which makes the whole time of prompt tuning longer. We leave the question ``How to accelerate the convergence of prompt tuning?'' to future work.

\subsection{Position of Prompt} \label{sec:prompt_pos}

\begin{figure}[t]
    \centering
    \includegraphics[width=\linewidth]{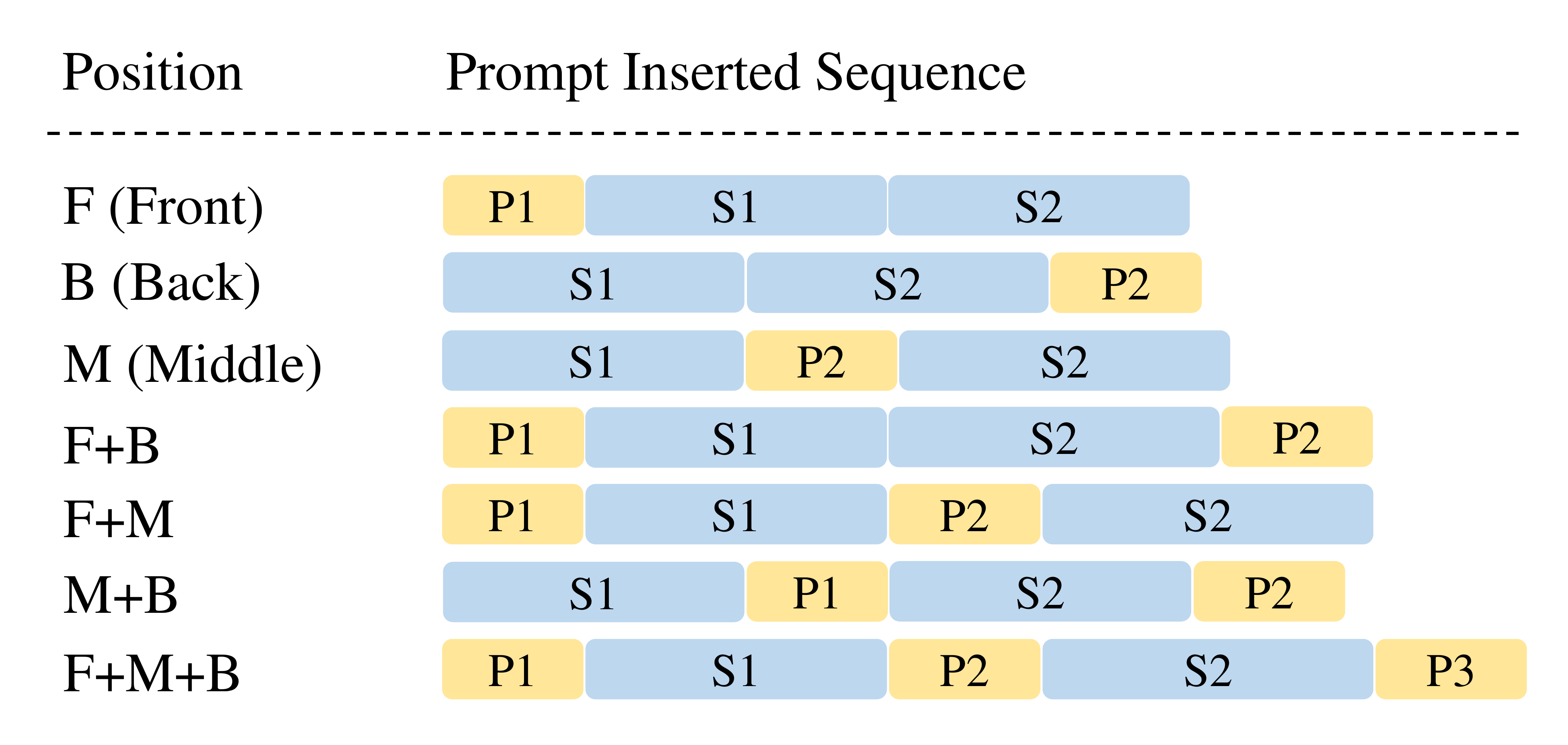}
    \caption{Different designs to insert prompts for the task with two input sequences, S1 and S2. P1, P2, P3 represent different input prompts. F, B, M represent Front, Back, and Middle, respectively.}
   \label{fig:p_insert}
\end{figure}

We study the effect of the positions of the inserted prompt tokens. For single-sequence tasks, such as Math23k, there exist $3$ strategies to insert the prompt: front, back, and front + back. For multi-sequence tasks, such as LCQMC, prompt tokens can also be inserted between two of the input sequences (middle). For a two-sequence input, there are $7$ strategies to insert the prompt tokens. The illustration of all possible prompt insertions of the two-sequence input task is shown in Figure~\ref{fig:p_insert}.

We conduct experiments on Math23k and LCQMC to evaluate the effect of prompt positions. We keep the number of prompt tokens as $100$. When there are $2$ positions to insert tokens, we insert $50$ tokens at each position. When there are $3$ positions, we insert $33$, $34$, $33$ tokens at each position. The results are shown in Table~\ref{tab:prompt_pos}.

\begin{table}[t]
    \centering
    \small
    \begin{tabular}{lrr}
    \toprule
           & Math23k & LCQMC \\
    \midrule
        F &  71.74   &   88.38   \\
        B &  72.40   &   88.50   \\
        F+B & 72.66 &  88.48   \\ \midrule
        M &    -     &  89.20       \\
        F+M &   -   &  90.21    \\
        M+B &  -    &  90.38    \\
        F+M+B & -  &  90.65      \\
    \bottomrule
    \end{tabular} 
    \caption{Effects of prompt positions on Math23k and LCQMC. For both datasets, we report the accuracy on dev sets.} \label{tab:prompt_pos}
    \vspace{-1em}
\end{table}

From the table, we have two observations. (1) For the single sentence task (Math23k), the positions of the prompt tokens have no significant influence on the model performance. (2) For the multi-sentence task (LCQMC), whether to insert the prompt between sentences significantly matters. Compared with inserting prompts between the two input sentences, only considering the front and the back positions leads to about $2\%$ accuracy drop. 

\begin{figure}[t]
    \centering
    \subfigure[Accuracy curve of different prompt positions.]{
        \centering
        \includegraphics[width=0.45\textwidth]{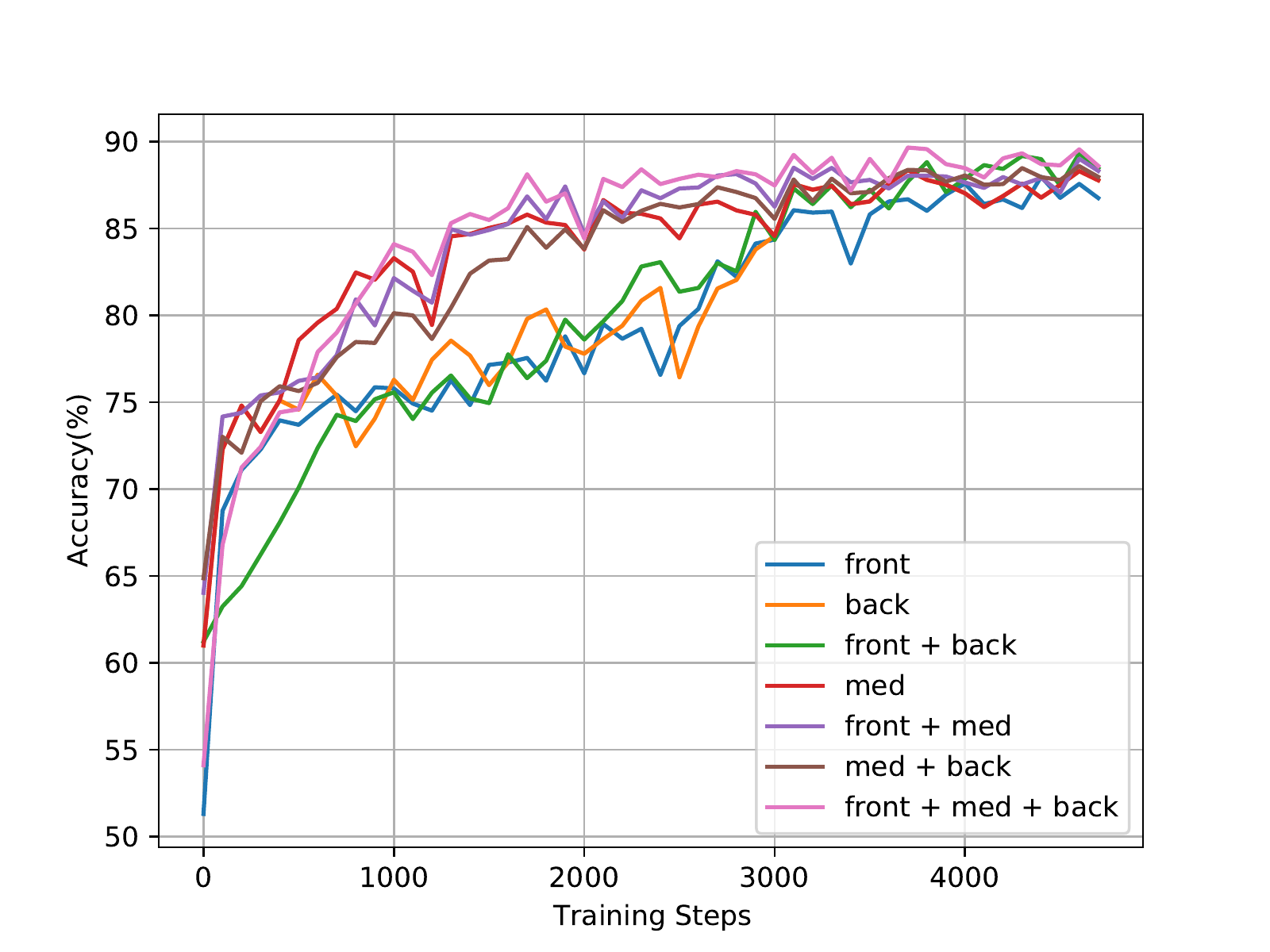}\label{fig:prompt_pos}
    }
    \subfigure[Accuracy curve of different ratio of the prompt token inserted between the two sentences.]{
        \centering
        \includegraphics[width=0.45\textwidth]{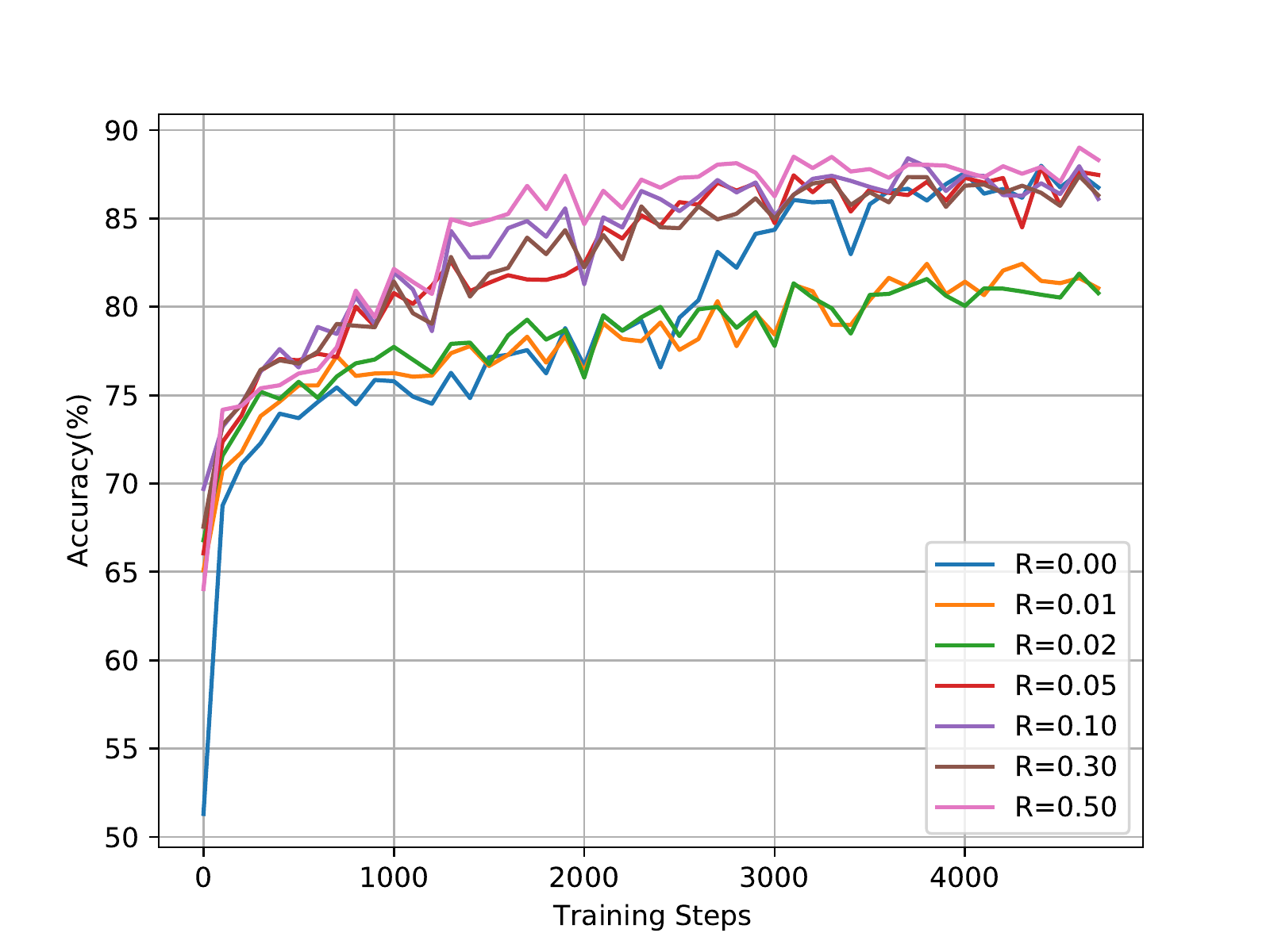}\label{fig:prompt_num}
    }
    \caption{Accuracy curves on the LCQMC dev set with different prompt insertion strategies.}
    \label{fig:prompt_pos_num}
\end{figure}

To study the effect in the learning process, we plot the accuracy curves on the LCQMC dev set of different prompt positions. Furthermore, we take F+M as an example and change the proportion of the number of prompt tokens at different positions. The results are shown in Figure~\ref{fig:prompt_pos_num}. In Figure~\ref{fig:prompt_num}, $R$ denotes the ratio between the middle prompt token number and the total prompt token number.

From the figure, we can conclude that: (1) As Figure~\ref{fig:prompt_pos} shows, for ``Front'', ``Back'' and ``Front + Back'' strategies, the convergence is much slower than the strategies with prompt tokens inserted between the two input sentences, which means it is necessary to insert the prompt between sentences to improve convergence speed. (2) As Figure~\ref{fig:prompt_num} shows, when $R=0.00$ (front)，$R=0.01$ and $R=0.02$ (insert $1$ or $2$ tokens between sentences), the model converges slowly.But when we insert $5$ or more tokens between the two sentences, the convergence speed is significantly improved. This means only a few middle-inserted tokens can help the model converge and when we add more tokens afterward, the impact of the middle token number is much less.

We think that the influence of the prompt token positions is related to the relative position embedding we use in CPM-2. When there are multiple input sentences, CPM-2 needs to model the tokens with a long distance. For relative position embedding, the long-range tokens will be assigned the same position embeddings, which may harm long-distance modeling. The prompt tokens inserted between sentences can bridge the gap between long-range tokens, which makes it easier for the model to learn the relationships between two input sentences.

\subsection{How Prompt Works} \label{sec:prompt_work}

Although prompt tuning can reach comparable performance with full-model tuning by only modifying a small number of parameters, the working mechanisms of prompt tuning are still unclear. We assume that the prompt can play two kinds of roles: (1) Working as a ``Provider''. Provide an additional context for the model input. (2) Working as an ``Aggregator''. Aggregate the information from the input text.

\begin{figure}[t]
    \centering
    \subfigure[The attention mask for ``Provider''. Attentions from prompt to text tokens are masked.]{
        \centering
        \includegraphics[width=0.42\textwidth]{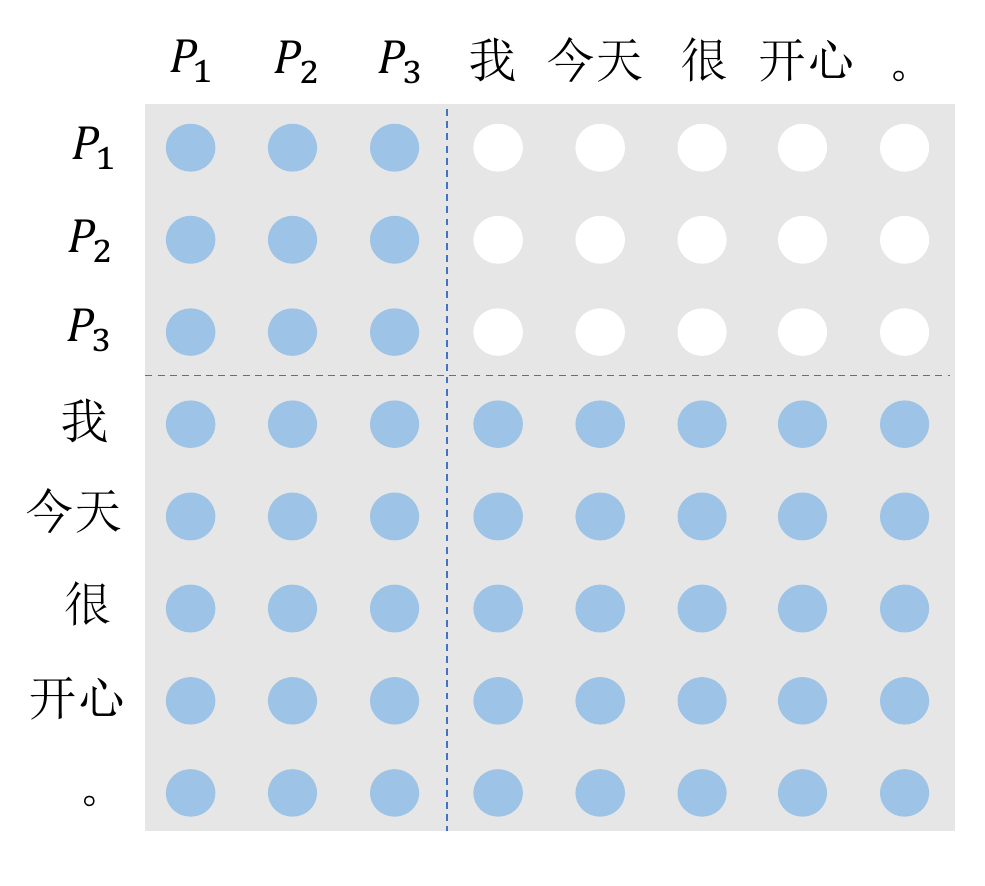}\label{fig:prompt_p2t}
    }
    \vspace{-1em}
    \subfigure[The attention mask for ``Aggregator''. Attentions from text tokens to prompt are masked.]{
        \centering
        \includegraphics[width=0.42\textwidth]{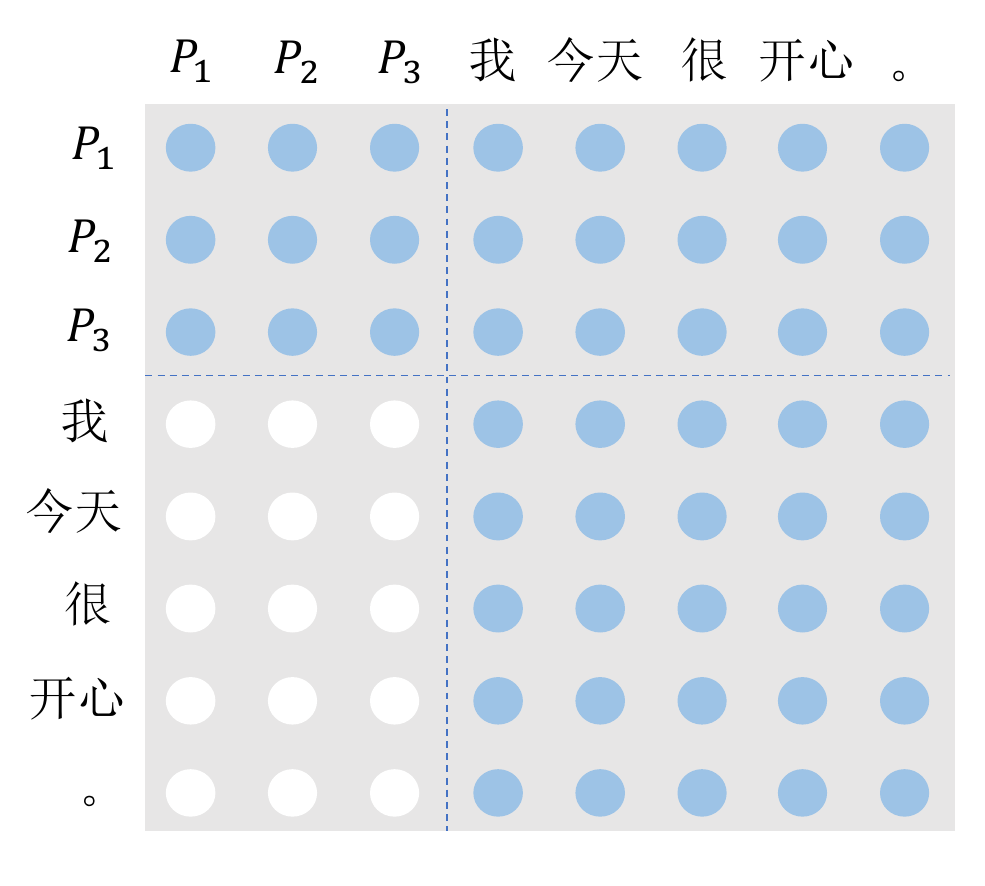}\label{fig:prompt_t2p}
    }
    \caption{Attention masks for "Provider" and "Aggregator". $P_1$, $P_2$, $P_3$ are prompt tokens.}
    \label{fig:prompt_mask}
    \vspace{-1em}
\end{figure}

To verify our hypothesis, we use attention masks to control the attentions between the prompt tokens and the text tokens. Specifically, for ``Provider'', we mask the attention from the prompt to text tokens such that the representations of prompt tokens can not be computed by attending to text tokens, disabling their ability to aggregate information. But they can still work as contexts and provide information to text tokens. For ``Aggregator'', on the contrary, we mask the attentions from text tokens to prompt tokens. In this way, prompt tokens can not work as contexts but can aggregate information by attending to text tokens. The illustration of our attention mask is shown in Figure~\ref{fig:prompt_mask}.

We add the attention masks mentioned above to the model when doing prompt tuning. We conduct experiments on on $\textrm{C}^3$, Math23k, LCQMC, and CCPM. The results are shown in Table~\ref{tab:mask_attn}.


\begin{table}[t]
    \centering
    \small
    \begin{tabular}{@{}lrrrr@{}}
    \toprule
                        & $\textrm{C}^3$& Math23K & LCQMC & CCPM\\ \midrule
    Full Attention    & 85.75 & 71.74   & 90.21 & 93.19\\
    Mask P to T & 83.84 & 69.92   & 81.50 & 92.78 \\
    Mask T to P & 68.54 & 35.29   & 79.45 & 86.90 \\ \bottomrule
    \end{tabular}
    \caption{Results of masking the attentions between prompts and texts. ``Mask P to T'' means masking the attention weights from the prompt to the text and ``Mask T to P'' means masking the attention weights from the text to the prompt. For both datasets, we report the accuracy on dev sets.}
    \label{tab:mask_attn}
    \end{table}

From the table, we can conclude that: (1) Both attention masks hurt the model performance on the two datasets. This means that the prompt should work as "Provider" and "Aggregator" at the same time for the model to reach good performance. (2) The impact of masking attention from text to prompt is larger than that of masking attention from prompt to text. This means prompt tokens are more likely to work as "Provider" than as "Aggregator" in prompt tuning.

\subsection{Two-Stage Fine-tuning} \label{sec:two_stage}

Previous work~\cite{schick2020exploiting,schick2020small} has shown that good prompts can help stimulate model ability in full-model tuning. However, most of them explore to manually design prompts or search prompts in a discrete space, which requires many human efforts. To make the prompt design easier, we attempt to search for good prompts in a continuous space, which can benefit full-model tuning afterward. Specifically, we propose to fine-tune models with two stages. In the first stage, we perform prompt tuning to search for a prompt suitable for the downstream task. Then, in the second stage, we fine-tune the whole model together with the prompt token embeddings. We hope that the model can take advantage of the prompt that we have found in the first stage and have better performance than the vanilla full-model tuning. We conduct experiments on $\textrm{C}^3$, Math23k, LCQMC, and CCPM. We try several prompts given by the first stage and select the one with the best results in the second stage. For each dataset, we use the same hyper-parameters as in Sections~\ref{sec:finetune} and~\ref{sec:prompt_main}. Our results on dev set are shown in Table~\ref{tab:two_stage_exp}.

\begin{table}[t]
\centering
\small
\begin{tabular}{@{}lrrrr@{}}
\toprule
          & $\textrm{C}^3$    & Math23k & LCQMC & CCPM\\ \midrule
CPM-2-F   & 85.66 & 73.85   & \textbf{90.88} & 93.00\\
CPM-2-P   & 85.75 & 71.74   & 90.21 & 93.19 \\
CPM-2-P+F & \textbf{86.77} & 75.26   & 90.45 & \textbf{93.42}\\
\ \ \ \ +fix prompt & 86.27     &   \textbf{76.17} & 89.64 & 93.01\\
\ \ \ \ -stage 1  &  85.04     &   72.40      &   88.76  & 92.76  \\ \bottomrule
\end{tabular}
\caption{Results of two-stage fine-tuning on three tasks using the dev sets. CPM-2-F stands for full-model tuning，CPM-2-P stands for prompt tuning. CPM-2-P+F is our two-stage fine-tuning. ``+fix prompt'' means we fix the parameters of the prompt we have found in stage 1 when we do full-model tuning in stage 2. ``-stage 1'' means we randomly initialize the prompt tokens and do full-model tuning directly without stage 1.}
\label{tab:two_stage_exp}
\end{table}

From the table, we have three observations: (1) Two-stage fine-tuning can significantly improve the model performance on $\textrm{C}^3$ and Math23k datasets by $2.16\%$ and $1.41\%$, respectively. On the LCQMC dataset, two-stage fine-tuning has a similar performance as vanilla full-model tuning. We think this is because the LCQMC dataset is relatively easier than the other two datasets and vanilla fine-tuning can perform well enough without a better prompt. (2) If we fix the prompt parameters during the second stage (``+fix prompt''), the model performance does not change much. We think this is because as fine-tuning goes, the gradients become small when backward to the input prompt. Therefore, the prompt tokens do not change much even when they are not fixed. (3) Without the first stage (``-stage  1''), even if we add additional parameters, the model can not reach a good performance, which proves the necessity of our two-stage fine-tuning.

\section{\INFSYS: Memory-Efficient Inference Framework for MoE Layers}

Although MoE linear layers could outperform dense linear layers with almost the same computational cost~\cite{fedus2021switch}, they greatly enlarge the number of model parameters and require more memory to store these parameters. When increasing the number of experts, the parameter size of the model can easily reach the order of tens or even hundreds of GBs. Such storage requirements greatly exceed the capacity of commodity GPUs, bringing difficulty not only to model training but also to model inference.

To make well-trained MoE layers more accessible to downstream tasks (e.g., to researchers using the aforementioned prompt tuning for downstream tasks), we introduce \INFSYS \footnote{\INFSYS is an open-source toolkit with MIT License at \url{https://github.com/TsinghuaAI/InfMoE}.}, a high-performance and memory-efficient inference framework that can offload parameters of experts of MoE layers to CPU memory.

\INFSYS enables the inference of MoE layers with hundreds of billions of parameters using one single GPU.
To preserve the efficiency of computation, we design a dynamic scheduling strategy that can overlap data movement of parameters with inference computation to the greatest extent.

\begin{figure}[t]
    \centering
    \subfigure[In na\"ive order: C2, C3, C4 must wait for loading.]{
        \centering
        \includegraphics[page=1,width=\linewidth]{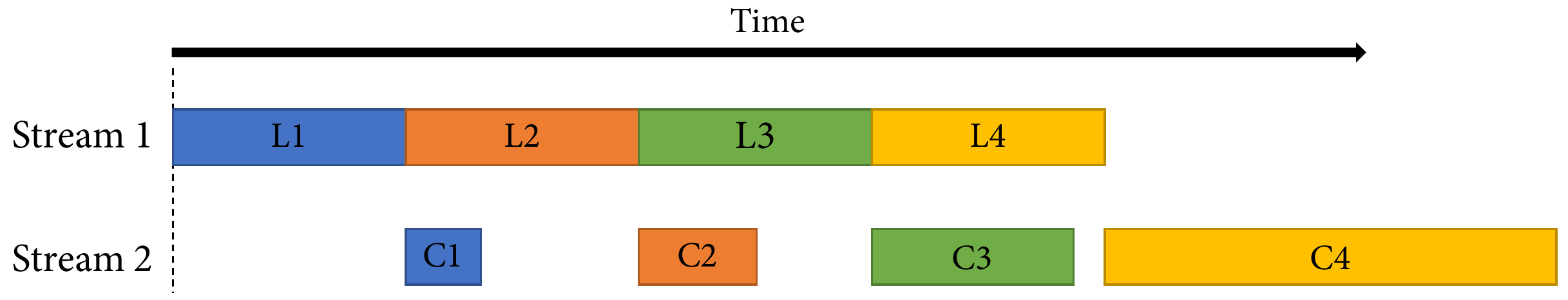}
        \label{fig:expert_sched:naive}
    }
    \subfigure[\INFSYS scheduling: all computation run without gaps.]{
        \centering
        \includegraphics[page=2,width=\linewidth]{Figures/expert_scheduling.pdf}
        \label{fig:expert_sched:dynamic}
    }
    \caption{Different scheduling strategies of load-imbalanced experts (L: parameter loading, C: computation).}
    \label{fig:expert_sched}
\end{figure}

\subsection{Existing Inference Frameworks}

PyTorch and TensorFlow are widely-used deep learning frameworks in industry and academia for both training and inference. There are also many other frameworks like TensorRT and ONNX Runtime that are specially designed for efficient model inference on different devices. However, they are currently not fit for the efficient inference of MoE layers for various reasons.

One category of these frameworks, like TensorFlow Serving, uses static computational graphs for training and inference. Typically, graphs can only be moved between CPUs and GPUs as a whole, so it is difficult to offload selected parameters in the inference process. Currently, no existing static-graph-based framework can provide full support for all required operators of MoE layers.

Another category, including PyTorch, uses dynamic computational graphs and provides simple interfaces to control data storage location (such as \texttt{layer.cuda()} and \texttt{layer.cpu()}. 
However, these frameworks usually take full control of the scheduling of computation and data movement. When handling MoE layers, they do not provide enough flexibility to implement the aforementioned overlapping mechanism. FastMoE~\cite{he2021fastmoe} is a novel high-performance MoE implementation on top of PyTorch. However, FastMoE focuses on large-scale distributed training and also lacks delicate control on scheduling.

TensorRT is a high-performance (yet relatively low-level) inference SDK developed by NVIDIA. It employs several optimization techniques like tensor fusion, kernel auto-tuning, and memory reusing. Our toolkit \INFSYS is developed based on TensorRT. The reason why we choose TensorRT is that it supports custom plugins. Therefore, we can implement our own plugin only for MoE layers with a specially designed scheduling strategy, handing over the remaining layers to TensorRT to get optimal performance. 


\subsection{Scheduling Strategy for Offloading}
The main challenge of the offloaded MoE layer inference lies in workload imbalance, as the amount of computation performed on different experts may be unbalanced. Tokens are routed and batched to different experts before computation. The workload distribution of experts may vary with different gating mechanisms \cite{lewis2021base, lepikhin2020gshard, fedus2021switch}. Experts having more tokens to process will spend more time in computation, while the overhead of data movement (which must be done prior to its computation) of each expert remains the same, for they all have the same amount of parameters.

In \INFSYS, by using different CUDA streams, parameter-loading and computation of different experts can be easily overlapped (i.e., executed at the same time). However, as shown in Figure~\ref{fig:expert_sched:naive}, na\"ively running experts in order easily leads to a waste of time on waiting for parameter loading due to the imbalanced computation time.

\begin{figure}[t]
\centering
\includegraphics[width = 1.0\linewidth]{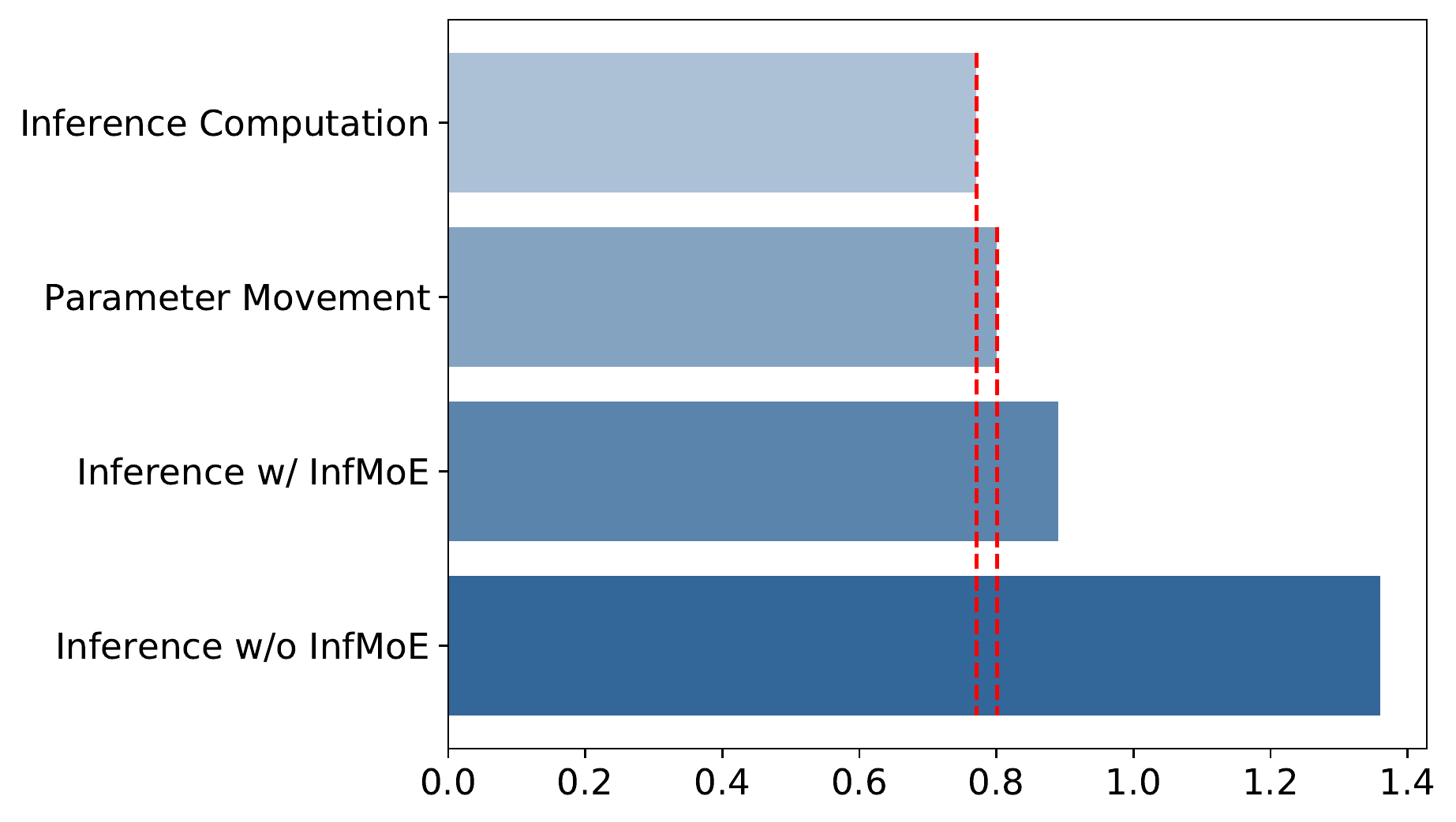}
\captionsetup{font={normalsize}}
\caption{For our MoE model with 32 experts, we give the time (seconds) of inference computation, parameter movement, inference with \INFSYS, and inference without \INFSYS.
}
\label{fig:informoetime}
\end{figure}

In order to maximize the overlap between the communication and computation, we design a dynamic schedule strategy in \INFSYS to reorder the loading and computation sequence of these experts:

Assuming there are $T$ experts in an MoE layer, we can estimate the computation time of the  $i$-th expert (denoted as $\alpha_{i}$) and its communication time (denoted as $\beta$).
$\alpha_i$ is obtained by dividing the number of floating operations by the peak computation performance of the GPU. With common expert workload (such as feed-forward layers in Transformers), it is proportional to the number of tokens. $\beta$ can be calculated as the size of parameters to load from the CPU divided by the peak bandwidth of the GPU. It remains the same for all experts.
In addition, due to the limit of GPU memory capacity and the existence of parameters belonging to non-MoE layers, only the parameters of a certain number (denoted as $K$ and can be either configured or automatically inferred) of experts can reside in GPU memory simultaneously.

In order to obtain optimal overlapping with negligible cost, \INFSYS use a greedy algorithm to generate a computation order of experts that satisfies the following two constraints:

\begin{itemize}
    \item $\forall 1 \le t \le T$, $\sum_{i=1}^{t-1} \alpha_{i} \ge (t - 1) \beta$. This means the parameter loading of each expert can be fully covered by the computation of previously loaded experts.
    \item $\forall 1 \le t \le T$, $\sum_{i=1}^{t-1} \alpha_{i} \le (t + K - 1) \beta$. This means no more than $K$ experts will be loaded to GPU memory simultaneously during the whole process.
\end{itemize}

This computation order can guarantee that no expert would have to wait for the loading of its parameters except the first one, thus fully hiding the overhead of data movement caused by offloading and leveraging full GPU computing performance (as shown in Figure \ref{fig:expert_sched:dynamic}. It is possible that these constraints cannot be satisfied at the same time. Such unsatisfiability indicates either the total computation amount is too small, or the workload is extremely imbalanced. The former cause can be mitigated by increasing the batch size, while the latter is out of the scope for inference. As for the MoE gating mechanism described in Section \ref{sec:pretraining}, it shows a relatively good balance between experts in our evaluation, thus fits well for \INFSYS. 

We evaluate the effectiveness of \INFSYS by inputting 40 instances into CPM-2-MoE with a single GPU. The computation times are reported in Figure~\ref{fig:informoetime}. From the figure, we can find that using \INFSYS for inference can overlap parameter movement and inference computation.

\section{More Promising Directions for Effective and Efficient Pre-trained Language Models}

In this section, we will briefly introduce our four novel explorations in tokenization, architecture, pre-training, and fine-tuning to achieve a more efficient pipeline of PLMs.

\subsection{Tokenization Based on Pronunciation and Glyph}

For Chinese PLMs, input tokenization is quite important. The conventional tokenization methods applied by existing PLMs may treat each character as an indivisible token. However, there is more linguistic information beyond characters. To explore a better tokenization method for Chinese PLMs, we consider pronunciation, glyph, and word segmentation to tokenize the input for PLMs. More specifically, we build pronunciation-based tokenizers, glyph-based tokenizers, and segmentation-based tokenizers respectively, and then 
systematically evaluate their performance based on BERT. Sufficient experimental results on various downstream NLU tasks have shown that applying pronunciation-based and glyph-based tokenizers can outperform existing used character-based tokenizers, and is more robust on the text noise. For more details, we refer to our paper~\citep{si2021shuowen}.

\subsection{Architecture Based on Non-Euclidean Geometry} 

Some recent efforts have shown that models learned in non-Euclidean geometry could better model complex data, especially those hyperbolic neural networks. However, existing hyperbolic networks are not completely hyperbolic, and training a deep hyperbolic network is also not trivial. To this end, we introduce a fully hyperbolic framework to build hyperbolic networks based on the Lorentz model and the Lorentz transformations. Based on the fully hyperbolic framework, we successfully train a hyperbolic Transformer and outperform existing Euclidean baselines. The experimental results show that hyperbolic Transformers can achieve comparable performance to Euclidean Transformers with half the size of model parameters, which may lead to more efficient PLMs in the future. In our paper~\citep{chen2021hybo}, we introduce more details of building hyperbolic neural networks.

\subsection{Pre-training Based on Knowledge Inheritance} 

As we mentioned before, large-scale PLMs have achieved success on various NLP tasks. However, training a large-scale PLM requires huge amounts of computational resources, which is time-consuming and expensive. Hence, taking the availability of existing well-trained PLMs into consideration is of importance. To this end, we propose knowledge inheritance to make previously trained PLMs benefit later larger PLMs. In fact, CPM-2 is built based on knowledge inheritance. In~\citep{qin2021ki}, we introduce the overall framework of knowledge inheritance, indicating the effect of teacher PLMs' settings, including pre-training methods, model architectures, training data, etc. For more details, we refer to our original paper.

\subsection{Fine-tuning Based on Rich Knowledge} 

In our experiments, we have shown that CPM-2 can perform well with prompt tuning, as additional prompts can stimulate the rich knowledge of PLMs to better serve downstream tasks. Besides model knowledge distributed in PLMs, we explore utilizing the prior knowledge to make fine-tuning PLMs more efficient and effective. To this end, we propose prompt tuning with rules, which can apply logic rules to construct prompts with several sub-prompts. By encoding prior knowledge of each class into prompt tuning, PLMs can converge faster and achieve better results on downstream tasks. More details of this part are included in our paper~\citep{han2021ptr}.

\section{Conclusion}

In this work, we propose a cost-effective pipeline for large-scale pre-trained language models, including pre-training with knowledge inheritance, fine-tuning based on prompt, and inference with dynamic scheduling. Correspondingly, we provide models and codes to support future applications with large-scale models. In the next stage, we will try to continually update our CPM models with emerging data gathered from the Internet to further improve model performance.

\section*{Acknowledgments}
Thanks to the Beijing Academy of Artificial Intelligence (BAAI) for providing the computing resources. In addition, we would like to thank BAAI, NetEase Inc., zhihu.com, and aminer.cn for the support in collecting the Chinese corpus.

\bibliography{anthology,custom}
\bibliographystyle{acl_natbib}

\appendix

\section{Contributions}

\noindent \textbf{Yuxian Gu and Zhengyan Zhang}
implemented the basic pre-training framework.

\vbox{}

\noindent \textbf{Xu Han} implemented the pipeline parallel strategy for better efficiency.

\vbox{}

\noindent \textbf{Zhengyan Zhang} implemented the MoE pre-training.

\vbox{}

\noindent \textbf{Yuxian Gu, Zhengyan Zhang, Chaojun Xiao, and Xu Han}
implemented the downstream tasks.

\vbox{}

\noindent \textbf{Shengqi Chen, Zhenbo Sun, Xu Han, and Yanzheng Cai}
implemented the toolkit of \INFSYS.

\vbox{}

\noindent \textbf{Jian Guan, Pei Ke, Guoyang Zeng, and Zhixing Tan}
prepared the pre-training data.

\vbox{}

\noindent \textbf{Yuan Yao and Fanchao Qi}
prepared the fine-tuning data.

\vbox{}

\noindent \textbf{Zhengyan Zhang, Yuxian Gu, Xu Han, Chaojun Xiao, Zhenbo Sun, and Shengqi Chen}
wrote the paper.

\vbox{}

\noindent \textbf{Zhiyuan Liu, Minlie Huang, and Wentao Han}
designed and led the research.

\vbox{}

\noindent \textbf{Yang Liu, Xiaoyan Zhu, Maosong Sun}
provided valuable advice to the research.

\end{CJK*}
\end{document}